\documentclass[letterpaper, 10 pt, conference]{ieeeconf}  

\usepackage[absolute, overlay]{textpos}
\textblockorigin{0mm}{3mm}

\usepackage{epsfig} 
\usepackage{amsmath}
\usepackage{amssymb}
\usepackage{caption}
\usepackage{array}
\usepackage{here}
\usepackage{cite}

\usepackage{graphicx}
\usepackage{subcaption}

\usepackage{tabularx}
\usepackage{multirow}
\usepackage{booktabs}
\usepackage[table,xcdraw]{xcolor}

\IEEEoverridecommandlockouts                              

\title{\LARGE \bf
Self-localization on a 3D map by fusing global and local features \\
from a monocular camera
}

\author{Satoshi Kikuchi$^{1}$ and Masaya Kato$^{2}$ and Tsuyoshi Tasaki$^{3}$
\thanks{*This work was not supported by any organization}
\thanks{$^{1}$Meijo University, 1-501 Shiogamaguchi, Tenpaku-ku, Nagoya, Aichi, Japan,
        {\tt\small 190442054@ccalumni.meijo-u.ac.jp}}%
\thanks{$^{2}$Meijo University, 1-501 Shiogamaguchi, Tenpaku-ku, Nagoya, Aichi, Japan,
{\tt\small 210442047@ccalumni.meijo-u.ac.jp}}%
\thanks{$^{3}$Meijo University, 1-501 Shiogamaguchi, Tenpaku-ku, Nagoya, Aichi, Japan,
        {\tt\small tasaki@meijo-u.ac.jp}}%
}

\begin{document}

\maketitle
\thispagestyle{empty}
\pagestyle{empty}

\begin{abstract}
Self-localization on a 3D map by using an inexpensive monocular camera is required to realize autonomous driving. 
Self-localization based on a camera often uses a convolutional neural network (CNN) that can extract local features that are calculated by nearby pixels. 
However, when dynamic obstacles, such as people, are present, CNN does not work well. 
This study proposes a new method combining CNN with Vision Transformer, which excels at extracting global features that show the relationship of patches on whole image. 
Experimental results showed that, compared to the state-of-the-art method (SOTA), the accuracy improvement rate in a CG dataset with dynamic obstacles is 1.5 times higher than that without dynamic obstacles. 
Moreover, the self-localization error of our method is 20.1\% smaller than that of SOTA on public datasets. 
Additionally, our robot using our method can localize itself with 7.51cm error on average, which is more accurate than SOTA. 
\end{abstract}

\begin{textblock}{19}(1,0)
        \copyright 2025 IEEE.  Personal use of this material is permitted.  Permission from IEEE must be obtained for all other uses, in any current or future media, including reprinting/republishing this material for advertising or promotional purposes, creating new collective works, for resale or redistribution to servers or lists, or reuse of any copyrighted component of this work in other works.

        Published in: 2025 IEEE/RSJ International Conference on Intelligent Robots and Systems (pp. 11043-11048)

\end{textblock}

\section{INTRODUCTION}
\begin{figure}[tb]
	\centering
	\includegraphics[width=\linewidth]{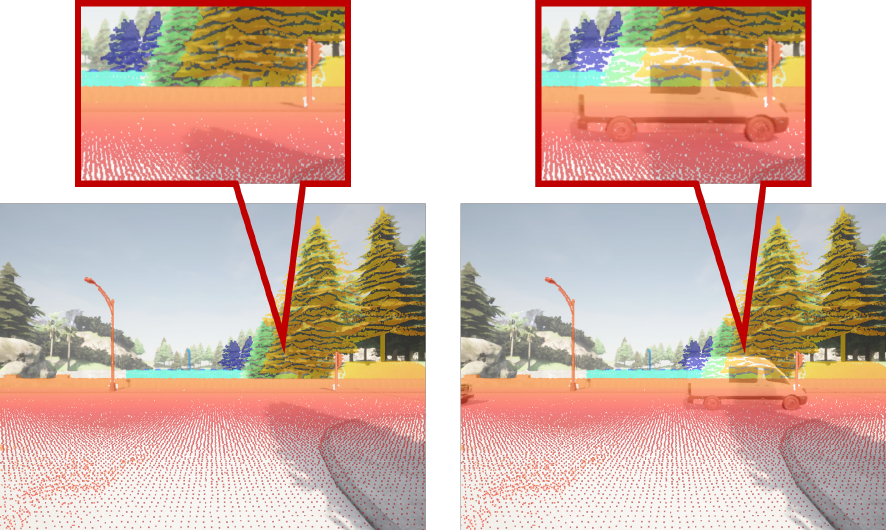}
	\caption{Local gap between color and depth images}
	\label{Fig:local_gap}
\end{figure}

\indent Self-localization in a 3D map is one of the key functions to realize autonomous driving. 
The 3D map is created using a vehicle equipped with multiple LiDARs. 
Here, we will define a 3D point cloud map that consists of point clouds from LiDARs as a 3D map. \\
\indent It is common for autonomous vehicles that use a LiDAR for self-localization. 
However, LiDARs are expensive, making it difficult to equip mass-market vehicles with them. 
Another common approach is to use GNSS for self-localization, but its accuracy decreases in places with many objects, such as urban areas. 
Therefore, some studies use a camera which is inexpensive and commonly equipped in mass-market vehicles \cite{cmrnet} \cite{lhmap-loc}. \\
\indent In camera-based self-localization, it is common to use color images obtained from a camera and depth image created by projecting a 3D map using rough self-pose as input. 
By inputting the color and depth image into a convolutional neural network (CNN), the precise self-pose on the 3D map can be estimated. 
Due to the nature of CNN, which performs convolution on nearby pixels, they are good at dealing with the relationships between nearby pixels as features. 
However, it is difficult to deal with the relationships between distant pixels. 
Here, the features that represent the relationships between nearby pixels are defined as local features. 
On the other hand, the features that represent relationships between distant pixels are defined as global features. \\
\indent Current CNN-based methods can estimate self-pose with high accuracy in static environments where dynamic obstacles, such as people and vehicles are absent. 
However, when dynamic obstacles are present, as shown in Fig. \ref{Fig:local_gap}, the input color and depth images have locally different areas. 
As a result, CNN, which relies on local features, has the problem of decreasing self-localization accuracy in dynamic environments with dynamic obstacles. 
Therefore, this study addresses the challenge of improving self-localization accuracy in environments with dynamic obstacles in order to achieve accurate self-localization using a camera. \\
\indent To address this challenge, we focused on Vision Transformer (ViT) \cite{vit}, which excels at extracting global features. 
Unlike CNN, ViT divides the whole image into patches and extracts features by calculating the similarity between all patches, making it possible to obtain global features. 
Therefore, by fusing the features obtained from CNN and ViT, we developed a new self-localization method that takes both global and local features into account. \\
\indent The academic contributions of this study are as follows: 

\begin{itemize}
	\item A new method that fuses global and local features was developed, and it was demonstrated that global features were effective in dynamic environments.
	\item It was demonstrated that our method was superior to the state-of-the-art method (LHMap-loc \cite{lhmap-loc}) in two public datasets: the KITTI dataset \cite{kitti} and the nuScenes dataset \cite{nuscenes}.
	\item Through driving experiments using a robot, it was demonstrated that our method estimated self-pose more accurately than LHMap-loc did.
\end{itemize}

\section{RELATED STUDIES}
\indent In this chapter, we introduce previous methods for obtaining precise self-pose on a 3D map using a camera and clarify the differences from our methods. 
Since this study does not focus on rough self-localization, methods that estimate precise self-pose based on rough self-pose are discussed as related studies. \\
\indent Self-localization using a 3D map and a camera can be classified into two types: mathematical optimization-based methods and end-to-end methods. 
Therefore, the following sections will explain both methods. 

\subsection{Mathematical optimization-based methods}
\indent Mathematical optimization-based methods often estimate self-pose by matching feature points. 
For example, Harris corner detector \cite{harris2d} and SuperPoint \cite{superpoint} are used for detecting 2D feature points from color images. 
For detecting 3D feature points from a 3D map, methods such as Harris 3D \cite{harris3d}, ISS \cite{iss}, and USIP \cite{usip} are utilized. 
Self-pose is estimated by minimizing the distance on the image between the 2D feature points and the projected 3D feature points using RANSAC \cite{ransac}. \\
\indent HF-Net \cite{hf-net} first estimates rough self-pose and then refines it for precise self-pose. 
Rough self-pose is estimated using a k-nearest neighbor search. 
Precise self-pose is computed through 2D-3D matching between feature points obtained from SuperPoint and a pre-generated 3D map using SfM (Structure from Motion). \\
\indent There are self-localization methods that minimize the distance between 3D point clouds rather than the distance between feature points on images. 
For example, the method \cite{3d-3d-reg} performs point cloud registration using ICP (Iterative Closest Point) \cite{icp} between a 3D map and 3D points generated by ORB-SLAM \cite{orb-slam}. \\
\indent Methods that utilize feature points often decrease accuracy due to the number of feature points. 
Therefore, this study adopts an end-to-end approach for self-localization. 

\subsection{End-to-End methods}
\indent In end-to-end methods, self-localization is primarily performed by inputting a color image obtained from a camera and a depth image into a neural network. 
The depth image is created by projecting a 3D map. 
CMRNet \cite{cmrnet} is the first self-localization method of the end-to-end approach. 
It is based on PWC-Net \cite{pwc-net}, which calculates optical flow, to establish the relationship between the 3D map and the color image to estimate self-pose. 
CMRNet extracts features from both color and depth images using CNN. \\
\indent LHMap-loc \cite{lhmap-loc} achieved the state-of-the-art (SOTA) on the KITTI dataset \cite{kitti} by using LHMap, which retains only the point clouds from the 3D map that are useful for self-localization. 
In LHMap-loc, the training process is divided into two stages: training for LHMap creation and training for self-localization using LHMap. 
In the LHMap creation training, a color image, an error-free depth image and an error-added depth image are used as inputs. 
The error-free depth image is created by projecting the 3D map using a ground-truth self-pose. 
The error-added depth image is created by projecting the 3D map using a rough self-pose. 
During training, the model learns to output the self-pose from the color image and the error-added depth image, while also generating a heatmap from the features of the error-free depth image. 
3D points projected onto the high value areas in the heatmap are selected to form the LHMap. 
In the self-localization training, an error-added depth image that is created by projecting the LHMap, and a color image are used as inputs. 
Features are extracted from each input using CNN, and the CNN is trained to output a precise self-pose. \\
\indent Current end-to-end methods often use CNN that is good at dealing with local features because it is easy to establish the relationship between the color and depth image. 
However, when there are large regions with local discrepancies between the color and depth images, accuracy decreases. 
To leverage global features, this study focuses on ViT \cite{vit}, which is commonly used in image recognition. 
Transformer-based methods are widely applied in image recognition \cite{sam} \cite{nddepth}, but there are no examples of their use in fusing local features for end-to-end self-localization method. 
This study develops a new method that fuses ViT features without negatively affecting the features obtained from CNN. 

\section{PROPOSED METHOD}
\begin{figure}[tb]
	\centering
	\includegraphics[width=\linewidth]{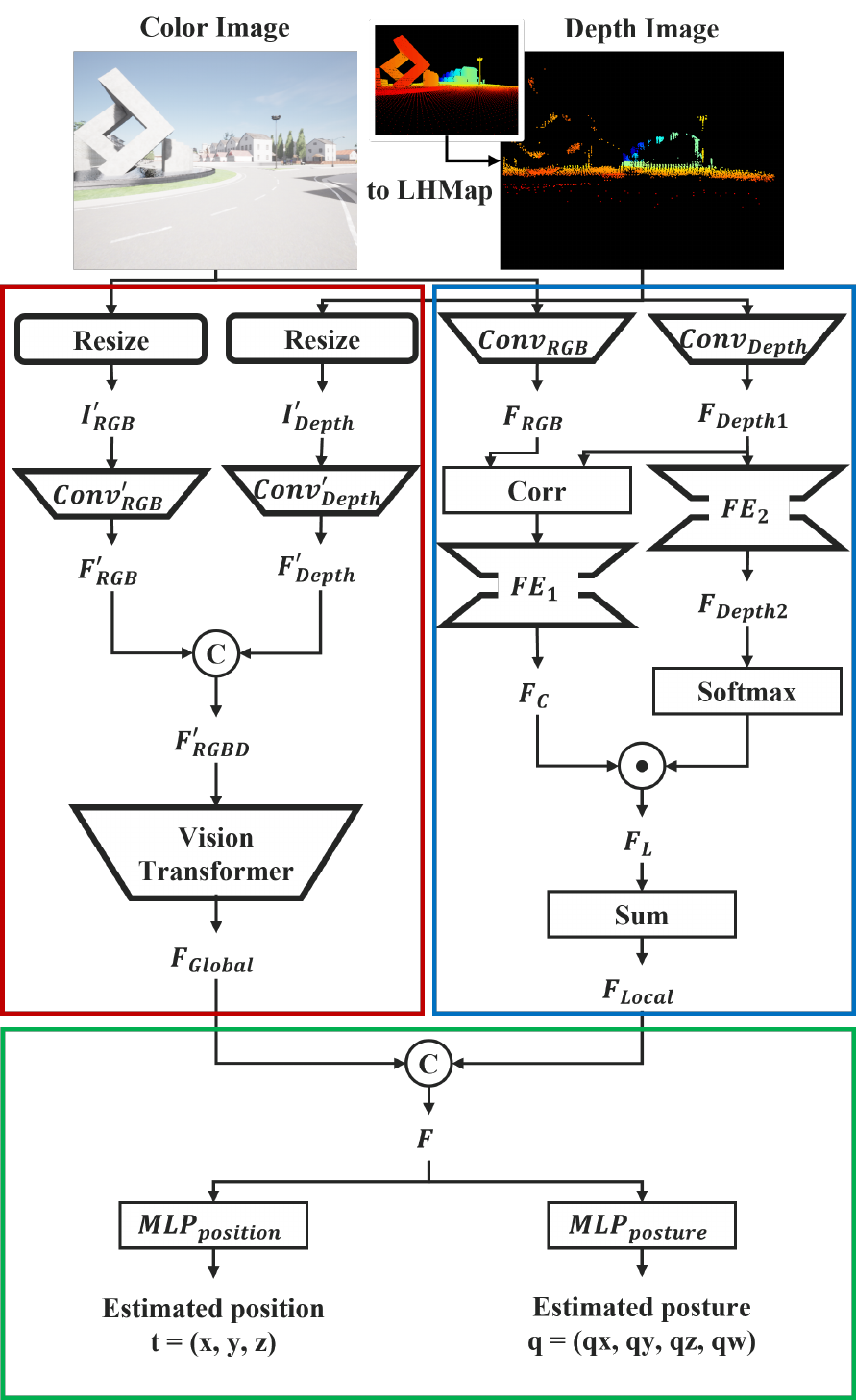}
	\caption{Overview of the proposed method}
	\label{Fig:network_overview}
\end{figure}

\indent The structure of the proposed neural network is shown in Fig. \ref{Fig:network_overview}. 
The proposed method takes a color image $I_{RGB} \in R^{3 \times h \times w}$ and a depth image $I_{Depth} \in R^{1 \times h \times w}$ as inputs to estimate the camera's position $t \in R^{3}$ and posture $q \in R^{4}$ on the 3D map.
Here, $I_{Depth}$ is created by LHMap that is made from a 3D map. 
The proposed method can be divided into three blocks: a global feature extraction block, a local feature extraction block, and a self-pose regression block. 
This chapter provides a detailed explanation of each block. 

\subsection{Global feature extraction block}
\indent The red box in Fig. \ref{Fig:network_overview} represents the global feature extraction block. 
This block utilizes ViT to learn the relationships between distant pixels. 
Since ViT requires a large amount of training data, this study uses a ViT model pre-trained on ImageNet \cite{imagenet} and fine-tunes it for the self-localization task. 
However, the features obtained by using pre-trained model as is affect negatively local features from CNN. 
Therefore, we use it after convolution for the color and depth image independently. 
Followings explain the structure for obtaining global feature in detail. \\
\indent As the pre-trained model has a fixed input tensor shape, preprocessing is applied to the images before feeding them into ViT. 
To obtain global features that do not affect negatively local features of both color and depth images, convolutional layers $Conv_{RGB}^{'}$ and $Conv_{Depth}^{'}$ are used to adjust the tensor shape. 
Specifically, $I_{RGB}$ and $I_{Depth}$ are first resized to $I_{RGB}^{'} \in R^{3 \times 447 \times 447}$ and $I_{Depth}^{'} \in R^{1 \times 447 \times 447}$ so that the outputs of $Conv_{RGB}^{'}$ and $Conv_{Depth}^{'}$ match the input size of the pre-trained ViT model. 
Then, by inputting $I_{RGB}^{'}$ and $I_{Depth}^{'}$ into $Conv_{RGB}^{'}$ and $Conv_{Depth}^{'}$ as shown in \eqref{eq:F_RGB^'} and \eqref{eq:F_Depth^'}, the tensors $F_{RGB}^{'} \in R^{2 \times 224 \times 224}$ and $F_{Depth}^{'} \in R^{1 \times 224 \times 224}$ are obtained. 

\begin{equation}
        F_{RGB}^{'}=Conv_{RGB}^{'}(I_{RGB}^{'}) \label{eq:F_RGB^'}
\end{equation}

\begin{equation}
        F_{Depth}^{'}=Conv_{Depth}^{'}(I_{Depth}^{'}) \label{eq:F_Depth^'}
\end{equation}

$F_{RGB}^{'}$ and $F_{Depth}^{'}$ are concatenated along the channel dimension as shown in \eqref{eq:F_RGBD^'} to obtain $F_{RGBD}^{'} \in R^{3 \times 224 \times 224}$. 
The obtained $F_{RGBD}^{'}$ is then used as the input to ViT. 

\begin{equation}
        F_{RGBD}^{'}=Concat(F_{RGB}^{'},F_{Depth}^{'} ) \label{eq:F_RGBD^'}
\end{equation}

The class token of the features output by ViT, with $F_{RGBD}^{'}$ as the input, is used as the global feature $F_{Global} \in R^{1 \times 768}$ in the self-pose regression block. 

\subsection{Local feature extraction block}
\indent The blue box in Fig. \ref{Fig:network_overview} represents the local feature extraction block. 
In this block, a CNN is used to learn the relationships between nearby pixels. 
The local feature extraction block utilizes the feature extraction module from LHMap-loc. 
The input images $I_{RGB}$ and $I_{Depth}$ are fed into convolutional layers $Conv_{RGB}$ and $Conv_{Depth}$, respectively, to obtain two feature maps $F_{RGB}$ and $F_{Depth}$ shown in \eqref{eq:F_RGB} and \eqref{eq:F_Depth_1}. 

\begin{equation}
        F_{RGB}=Conv_{RGB}(I_{RGB}) \label{eq:F_RGB}
\end{equation}

\begin{equation}
        F_{Depth_{1}}=Conv_{Depth}(I_{Depth}) \label{eq:F_Depth_1}
\end{equation}

$F_{RGB}$ and $F_{Depth_{1}}$ are input both the correlation calculation module $Corr$ and the feature extractor $FE_{1}$ which consists of convolutional and deconvolutional layers. 
Then, a correlation map $F_{C}$ representing the relationship between $I_{RGB}$ and $I_{Depth}$ is obtained as shown in \eqref{eq:F_C}. 

\begin{equation}
        F_{C}=FE_{1}(Corr(F_{RGB}, F_{Depth_{1}})) \label{eq:F_C}
\end{equation}

Furthermore, $F_{Depth_{1}}$ is passed through the feature extractor $FE_{2}$, which consists of convolutional and deconvolutional layers, to calculate the feature map $F_{Depth_{2}}$ as shown in \eqref{eq:F_Depth_2}. 

\begin{equation}
        F_{Depth_{2}}=FE_{2}(F_{Depth_{1}}) \label{eq:F_Depth_2}
\end{equation}

By inputting $F_{Depth_{2}}$ the softmax layer and calculating the Hadamard product with $F_{C}$ as shown in \eqref{eq:F_L}, the feature map $F_{L} \in R^{512 \times h^{'} \times w^{'}}$ is obtained. 
Here, $h^{'}=h/16$ and $w^{'}=w/16$. 

\begin{equation}
        F_{L}=F_{C} \bigodot Softmax(F_{Depth_{2}}) \label{eq:F_L}
\end{equation}

Finally, by summing up the elements of each channel in $F_{L}$ as shown in \eqref{eq:F_Local}, the local feature $F_{Local} \in R^{1 \times 512}$ is obtained. 

\begin{equation}
        F_{Local}=\sum_{h^{'} \times w^{'}}F_{L} \label{eq:F_Local}
\end{equation}

\subsection{Self-pose regression block}
\indent The green box in Fig. \ref{Fig:network_overview} represents the self-pose regression block. 
In this block, the self-pose $t$ and $q$ are estimated using the local feature $F_{Local}$ and the global feature $F_{Global}$. 
First, $F_{Local}$ and $F_{Global}$ are concatenated as shown in \eqref{eq:F_Concat} to obtain the fused feature $F \in R^{1 \times 1280}$. 

\begin{equation}
        F=Concat(F_{Local}, F_{Global}) \label{eq:F_Concat}
\end{equation}

By inputting the fused feature $F$ into $MLP_{position}$ and $MLP_{posture}$, the self-position and self-posture are estimated as shown in \eqref{eq:MLP_position} and \eqref{eq:MLP_posture}. 
$MLP_{position}$ and $MLP_{posture}$ consist of three fully connected layers with activation functions, and their structure is the same as LHMap-loc, except for the input shape of the first layer. 

\begin{equation}
        t=MLP_{position}(F) \label{eq:MLP_position}
\end{equation}

\begin{equation}
        q=MLP_{posture}(F) \label{eq:MLP_posture}
\end{equation}

\section{EVALUATION EXPERIMENT}
\subsection{Experiment Overview}
\indent To compare the accuracy of the proposed method in static and dynamic environments, as well as to evaluate its accuracy against previous methods, the following three experiments were conducted. 

\begin{enumerate}
	\item Accuracy comparison in static and dynamic environments using a CG dataset
	\item Accuracy comparison on a public dataset
	\item Accuracy comparison using a mobile robot
\end{enumerate}

\indent In these experiments, LHMap-loc \cite{lhmap-loc}, which achieved SOTA on the KITTI dataset \cite{kitti} among CNN-based methods, is used as the previous method. 
ViT used for feature extraction in the proposed method uses a pre-trained model developed by Google. 
The evaluation metrics are the mean [cm] and median [cm] of self-localization errors in each dataset.

\subsection{Datasets}
\begin{figure}[tb]
        \centering
        \begin{subfigure}{0.49\linewidth}
                \centering 
                \includegraphics[width=\linewidth]{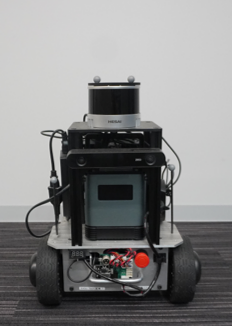}
                \caption{Robot used for mapping}
                \label{Fig:map-robot}
        \end{subfigure}
        \hfill
        \begin{subfigure}{0.49\linewidth}
                \centering 
                \includegraphics[width=\linewidth]{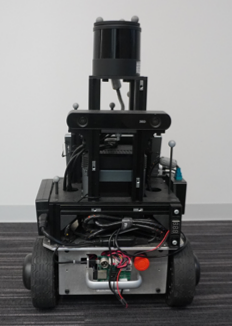}
                \caption{Robot used in the evaluation}
                \label{Fig:loc-robot}
        \end{subfigure}
        \caption{Robot used in the experiment}
        \label{Fig:robot}
\end{figure}

\indent In Experiment 1, two types of CG data, static and dynamic environments, are used, created from seven different towns in the CARLA simulator \cite{carla}. 
The resolution of the color images is set to 640px in height and 832px in width. 
For the training data, a total of 3500 images are used, with 500 images from the beginning of each town. 
For the validation data, 700 images are used, consisting of 100 images following training data of each town. 
The evaluation data consists of 7000 images, with 1000 images from each town that were not used in either the training or validation data. 
The same trajectory data is used for both static and dynamic environments. 
In the dynamic environment, from 30 to 70 people and vehicles were placed as dynamic obstacles in each town. \\
\indent In Experiment 2, two public datasets, KITTI \cite{kitti} and nuScenes \cite{nuscenes} are used. 
The KITTI dataset consists of 11 sequences collected from driving in Germany. 
The nuScenes dataset consists of a total of 4 towns, including 3 towns in Singapore and 1 town in the United States. \\
\indent For the KITTI dataset, 7 sequences were used to match the experimental setup of LHMap-loc. 
In the KITTI dataset, for the training data, a total of 3500 images are used, with 500 images from the beginning of each town. 
For the validation data, 700 images are used, consisting of 100 images from the continuation of each town. 
The evaluation data consists of the remaining 11766 images, that were not used in either the training or validation data. 
For input image size adjustment, zero-padding is applied to the bottom right of the color images so that the resolution becomes 384px in height and 1280px in width. \\
\indent For the nuScenes dataset, we used 3 towns in Singapore. 
Since the nuScenes dataset has significant differences in the number of data samples across towns, the data was split with weighting applied to each town accordingly. 
For the training data, a total of 3477 images were used, with 363 images from Holland Village, 1555 images from One North, and 1559 images from Queenstown, all collected from the beginning of each town. 
For the validation data, a total of 677 images were used, with 40, 317, and 320 images following training data from each town. 
For the evaluation data, a total of 7223 images were used, with 367, 5436, and 1420 daytime images from each town. 
The evaluation data are not used for training or validation. \\
\indent For the nuScenes dataset, the color images are resized by a scale factor of 3/4 and then cropped based on the image center to achieve a final resolution of 640px in height and 832px in width. \\
\indent In Experiment 3, the Meijo University dataset, collected from within the Meijo University campus, is used. 
This dataset consists of a 3D map, color images, and ground-truth self-pose. 
The 3D map was created using a robot equipped with a HESAI XT32 LiDAR, as shown in Fig. \ref{Fig:map-robot} in autumn. 
The color images and ground-truth self-pose were obtained using a robot equipped with a HESAI QT128 LiDAR and a Stereolabs ZED2 stereo camera, as shown in Fig. \ref{Fig:loc-robot} in winter. 
Only the left camera images from the stereo camera were used. 
The images, originally at a resolution of 1080px in height and 1920px. 
In the experiment, we resized them by a scale factor of 2/3, then cropped based on the image center to achieve a final resolution of 640px in height and 832px in width. 
The ground-truth self-pose was obtained using NDT scan matching \cite{ndt-matching}, which is implemented in Autoware \cite{autoware}. 
In the Meijo University dataset, the number of images for the training, validation and evaluation data are 4949, 1651, and 32993, respectively. \\
\indent For evaluation of all datasets, the rough self-pose used for correction was generated by adding a random error of -60cm to 60cm to the ground-truth self-pose, which is the same as LHMap-loc. 

\subsection{Experimental results and discussion}
\begin{table}[h]
        \centering
        \caption{Quantitative Result (CARLA)}
        \label{tab:result_carla}
        \vspace{-5pt}
        \begin{tabular}{c|c|cc}
                \hline
                                              &                          & \multicolumn{2}{c}{Position error $[$cm$]$}                                                        \\ \cline{3-4} 
                \multirow{-2}{*}{Environment} & \multirow{-2}{*}{Method} & \multicolumn{1}{c|}{Mean}                                  & Median                                \\ \hline
                                              & Previous                 & \multicolumn{1}{c|}{15.35}                                 & 13.76                                 \\ \cline{2-4} 
                \multirow{-2}{*}{Static}      & Proposed                 & \multicolumn{1}{c|}{{\color[HTML]{CB0000} \textbf{14.89}}} & {\color[HTML]{CB0000} \textbf{13.42}} \\ \hline
                                              & Previous                 & \multicolumn{1}{c|}{16.00}                                 & 14.90                                 \\ \cline{2-4} 
                \multirow{-2}{*}{Dynamic}     & Proposed                 & \multicolumn{1}{c|}{{\color[HTML]{CB0000} \textbf{15.27}}} & {\color[HTML]{CB0000} \textbf{13.90}} \\ \hline
        \end{tabular}
        \vspace{2pt}
        \caption{Quantitative Result (Public dataset)}
        \label{tab:result_public}
        \vspace{-5pt}
        \begin{tabular}{c|c|cc}
                \hline
                                           &                          & \multicolumn{2}{c}{Position error $[$cm$]$}                                                      \\ \cline{3-4} 
                \multirow{-2}{*}{Dataset}  & \multirow{-2}{*}{Method} & \multicolumn{1}{c|}{Mean}                                 & Median                               \\ \hline
                                           & Previous                 & \multicolumn{1}{c|}{9.58}                                 & 8.68                                 \\ \cline{2-4} 
                \multirow{-2}{*}{KITTI}    & Proposed                 & \multicolumn{1}{c|}{{\color[HTML]{CB0000} \textbf{9.36}}} & {\color[HTML]{CB0000} \textbf{8.31}} \\ \hline
                                           & Previous                 & \multicolumn{1}{c|}{8.81}                                 & 7.32                                 \\ \cline{2-4} 
                \multirow{-2}{*}{nuScenes} & Proposed                 & \multicolumn{1}{c|}{{\color[HTML]{CB0000} \textbf{7.04}}} & {\color[HTML]{CB0000} \textbf{5.91}} \\ \hline
        \end{tabular}
        \vspace{2pt}

        \caption{Quantitative Result (Meijo University)}
        \label{tab:result_meijo}
        \vspace{-5pt}
        \begin{tabular}{c|c|cc}
                \hline
                                          &                          & \multicolumn{2}{c}{Position error $[$cm$]$}                                                      \\ \cline{3-4} 
                \multirow{-2}{*}{Dataset} & \multirow{-2}{*}{Method} & \multicolumn{1}{c|}{Mean}                                 & Median                               \\ \hline
                                          & Previous                 & \multicolumn{1}{c|}{8.18}                                 & 6.81                                 \\ \cline{2-4} 
                \multirow{-2}{*}{Meijo}   & Proposed                 & \multicolumn{1}{c|}{{\color[HTML]{CB0000} \textbf{7.51}}} & {\color[HTML]{CB0000} \textbf{6.30}} \\ \hline
        \end{tabular}
\end{table}

\begin{figure}[h]
        \centering
        \includegraphics[width=\linewidth]{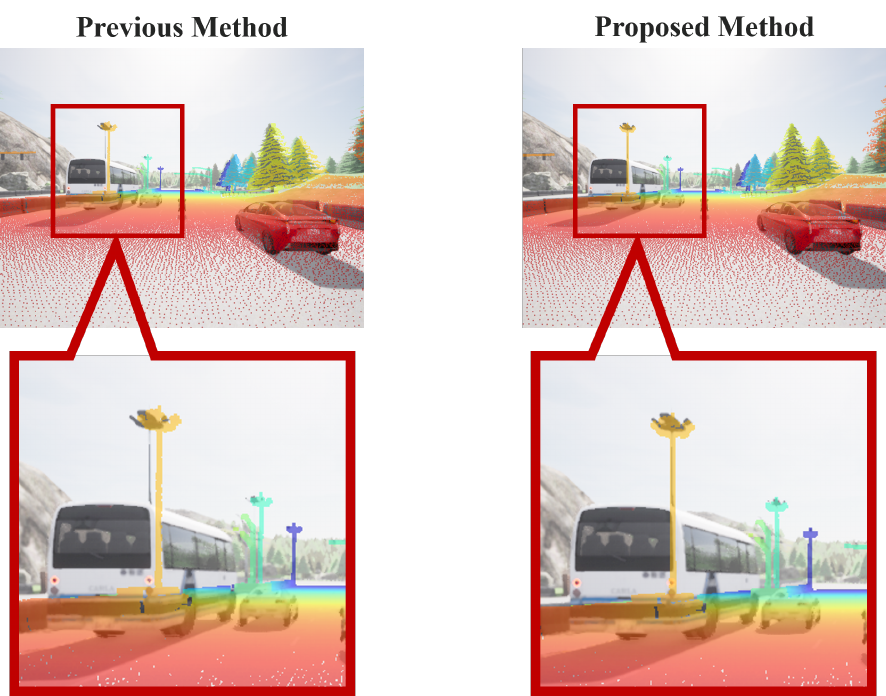}
        \caption{Qualitative Result (CARLA / Dynamic)}
        \label{Fig:qualitative_carla}
\end{figure}

\begin{figure}[tb]
        \centering
        \includegraphics[width=\linewidth]{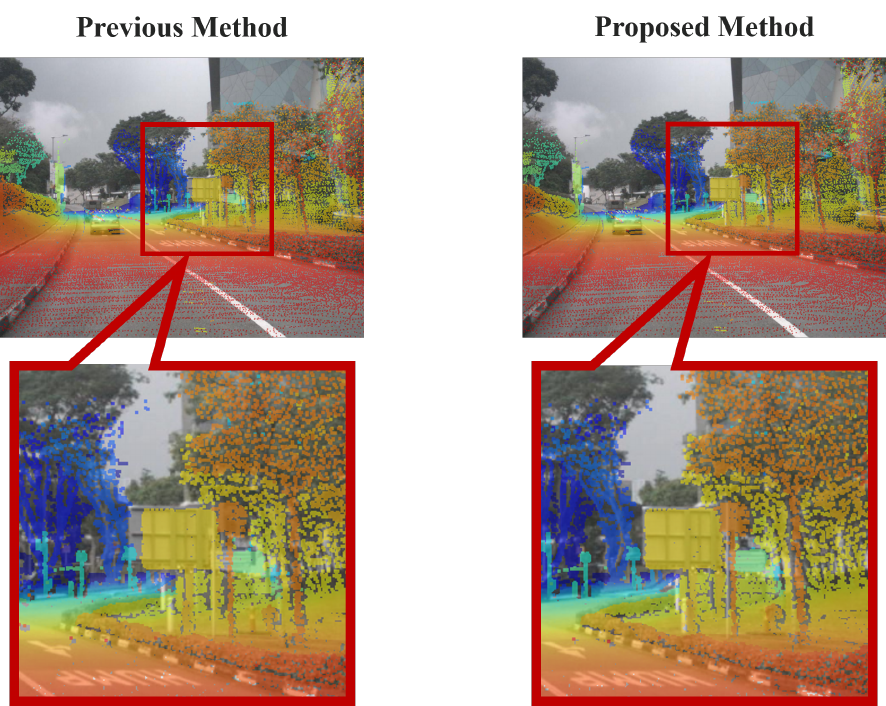}
        \caption{Qualitative Result (nuScenes)}
        \label{Fig:qualitative_nuscenes}

        \centering
        \includegraphics[width=\linewidth]{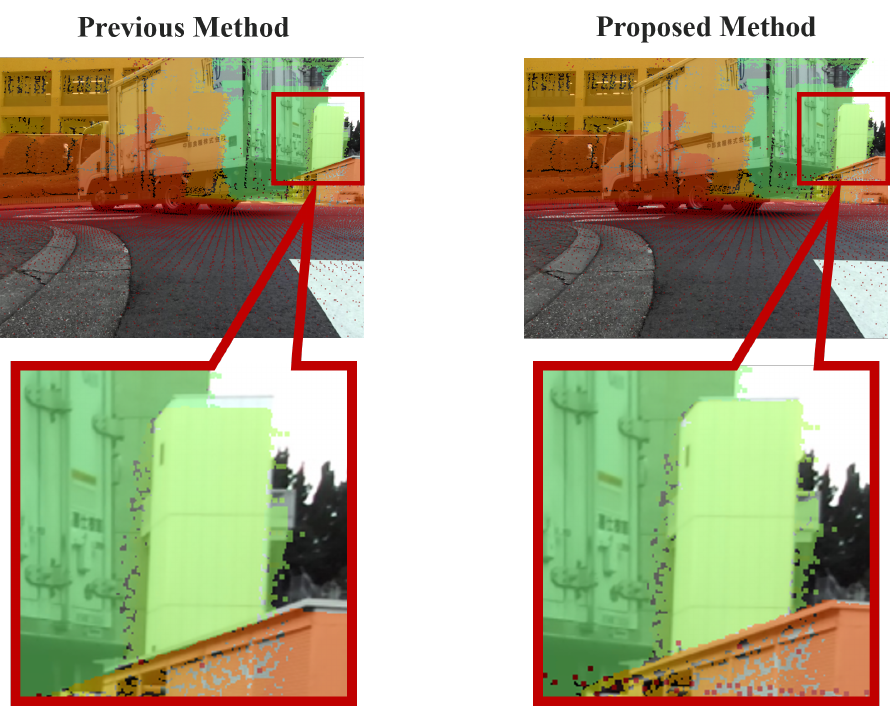}
        \caption{Qualitative Result (Meijo University)}
        \label{Fig:qualitative_meijo}
\end{figure}

\indent The self-localization errors for each method in Experiments 1, 2, and 3 are shown in Table \ref{tab:result_carla}, Table \ref{tab:result_public}, and Table \ref{tab:result_meijo}, respectively. 
Fig. \ref{Fig:qualitative_carla}, Fig. \ref{Fig:qualitative_nuscenes}, and Fig. \ref{Fig:qualitative_meijo} show the images created by projecting the 3D map based on the self-pose estimated by each method. \\
\indent We compare the results of the static environment with those of dynamic environments. 
As shown in Table \ref{tab:result_carla}, the previous method showed a 4.2\% increase in error in the dynamic environment, whereas the proposed method limited the error increase to 2.6\%. 
Additionally, the error of our method is 3.0\% smaller than that of LHMap-loc in the static environment, and 4.6\% smaller in the dynamic environment. 
Therefore, accuracy improvement rate in the dynamic environment is 1.5 times greater than that in the static environment. 
Therefore, it is confirmed that extracting global features with ViT enhances self-localization accuracy, particularly in scenes with dynamic obstacles. 
Certainly, as shown in Fig. \ref{Fig:qualitative_carla}, the projected 3D map overlaps a color image by using the pose estimated by our method. 
In the scene shown in Fig. \ref{Fig:qualitative_carla}, the previous method resulted in a self-localization error of 31.6cm, whereas our method reduced the error to 7.5cm. \\
\indent As shown in Table \ref{tab:result_public}, self-localization errors were decreased on both public datasets. 
Especially on the nuScenes dataset, the error was reduced by 20.1\% compared to the previous method. 
This demonstrates that the proposed method is effective not only in CG environments but also in real-world urban environments. \\
\indent As shown in Table \ref{tab:result_meijo}, our method achieved an 8.2\% reduction in error compared to the previous method on the Meijo University dataset. 
Despite using a 3D map created at a different time when the color images were captured, accurate self-localization was achieved. 
Since real-world environments continuously change over time, local discrepancies between the map and images are likely to occur, in addition to the presence of dynamic obstacles. 
Therefore, fusing global features using ViT is also beneficial for a map management. 

\section{ABLATION STUDY}
\indent A comparative experiment is conducted on how the color and depth images are input into ViT. \\
\indent This experiment compares the following four methods:

\begin{enumerate}
	\item Proposed method
	\item Resize only the color image (without using depth)
	\item Resize the color image and apply convolution (without using depth)
	\item Concatenate the color and depth image along the channel dimension to create an RGBD image, then apply resizing and convolution
\end{enumerate}

\indent In this experiment, the nuScenes dataset is used. 
The results are shown in Table. \ref{tab:result_ablation}. 
As shown in Table. \ref{tab:result_ablation}, our method achieves the smallest error in self-localization among all methods. 
The error of Method 2 is much larger than that of Method 3. 
One reason why Method 2 shows large error is the lack of convolution. 
When the tensor shape is adjusted using only resizing, the resolution must be reduced to 224x224. 
However, when convolution is applied after resizing, the intermediate resolution remains at 447x447, preserving more information. 
Additionally, convolution allows for extracting global features while considering local features. 
Thus, it can be inferred that global features were successfully fused without negatively affecting local features. \\
\indent Furthermore, in the proposed method, convolution is applied separately to both the color and depth images. 
Therefore, the tensor input to ViT in our method has depth information more strongly than the tensor in Method 4. 
The effective utilization of depth information is considered to be another factor contributing to the higher accuracy of our method in self-localization.

\begin{table}[tb]
        \centering
        \caption{Quantitative Result (Ablation study)}
        \label{tab:result_ablation}
        \begin{tabular}{c|c|cc}
                \hline
                                           &                          & \multicolumn{2}{c}{Position error $[$cm$]$}                                                      \\ \cline{3-4} 
                \multirow{-2}{*}{Dataset}  & \multirow{-2}{*}{Method} & \multicolumn{1}{c|}{Mean}                                 & Median                               \\ \hline
                                           & Proposed                 & \multicolumn{1}{c|}{{\color[HTML]{CB0000} \textbf{7.04}}} & {\color[HTML]{CB0000} \textbf{5.91}} \\ \cline{2-4} 
                                           & RGB / resize             & \multicolumn{1}{c|}{14.15}                                & 12.36                                \\ \cline{2-4} 
                                           & RGB / resize + conv      & \multicolumn{1}{c|}{7.13}                                 & {\color[HTML]{CB0000} \textbf{5.91}} \\ \cline{2-4} 
                \multirow{-4}{*}{nuScenes} & RGBD / resize + conv     & \multicolumn{1}{c|}{8.27}                                 & 6.97                                 \\ \hline
        \end{tabular}
\end{table}

\section{CONCLUSION}
\indent This study aimed to achieve self-localization using a monocular camera, addressing the challenge of improving self-localization in environments with dynamic obstacles. 
To address this challenge, a new method was developed that fuses CNN and ViT, obtaining both global and local features from images. 
Experimental results showed that our method could achieve accurate self-localization even in environments with dynamic obstacles. 
Additionally, in experiments using public datasets and a mobile robot, the self-localization error was reduced by 20.1\% compared to the SOTA method.\\
\indent This study successfully achieved self-localization generally suitable for autonomous driving in average. 
However, occasional large errors still occur. 
In future work, particle filters and other techniques will be used to further stabilize self-localization, aiming to demonstrate its feasibility for autonomous driving. 


\end{document}